\documentclass[10pt,twocolumn,letterpaper]{article}

\usepackage{cvpr}
\usepackage{times}
\usepackage{epsfig}
\usepackage{graphicx}
\usepackage{amsmath}
\usepackage{amssymb}
\usepackage{xcolor}
\usepackage{blindtext}
\usepackage{overpic}
\usepackage{transparent}
\usepackage{pifont}
\usepackage{booktabs}
\usepackage{multirow}
\usepackage{tabularx}
\usepackage[super]{nth}
\usepackage[format=plain,labelformat=simple,labelsep=period,font=small,skip=4pt,compatibility=false]{caption}
\usepackage[font=footnotesize,skip=2pt]{subcaption}

\makeatletter
\newcommand\notsotiny{\@setfontsize\notsotiny\@vipt\@viipt}
\newcommand{\myparagraph}[1]{\medbreak\noindent\textbf{#1}}
\newcommand{\cmark}{\ding{51}}

\makeatother

\usepackage{etoolbox}
\usepackage[binary-units]{siunitx}
\robustify\bfseries
\DeclareSIUnit{\inch}{inch}

\usepackage[pagebackref=true,breaklinks=true,letterpaper=true,colorlinks,bookmarks=false]{hyperref}
\usepackage[capitalise]{cleveref}
\cvprfinalcopy

\def\cvprPaperID{714}

\ifcvprfinal\fi
\begin{document}

\newcolumntype{C}[1]{>{\centering\arraybackslash}p{#1}}

\title{Learning by Analogy: Reliable Supervision from Transformations \\
for Unsupervised Optical Flow Estimation}

\author{Liang Liu$^1$\thanks{Work mainly done during an internship at Tencent Youtu Lab.} ~ ~ Jiangning Zhang$^1$ ~ ~ Ruifei He$^1$ ~ ~ Yong Liu$^1$\thanks{Corresponding author.} ~ ~ Yabiao Wang$^2$ \\ Ying Tai$^2$ ~ Donghao Luo$^2$ ~ ~ Chengjie Wang$^2$ ~ ~ Jilin Li$^2$ ~ ~ Feiyue Huang$^2$ \\
\normalsize $^1$ Zhejiang University ~ ~ $^2$Youtu Lab, Tencent \\
{\tt\small \{leonliuz, 186368, rfhe\}@zju.edu.cn, yongliu@iipc.zju.edu.cn} \\
{\tt\small \{caseywang, yingtai, michaelluo, jasoncjwang, jerolinli, garyhuang\}@tencent.com}
}

\maketitle
\thispagestyle{empty}

\begin{abstract}
Unsupervised learning of optical flow, which leverages the supervision from view synthesis, has emerged as a promising alternative to supervised methods.
However, the objective of unsupervised learning is likely to be unreliable in challenging scenes.
In this work, we present a framework to use more reliable supervision from transformations. It simply twists the general unsupervised learning pipeline by running another forward pass with transformed data from augmentation, along with using transformed predictions of original data as the self-supervision signal. Besides, we further introduce a lightweight network with multiple frames by a highly-shared flow decoder. Our method consistently gets a leap of performance on several benchmarks with the best accuracy among deep unsupervised methods. Also, our method achieves competitive results to recent fully supervised methods while with much fewer parameters.
\vspace{-1em}
\end{abstract}

\section{Introduction}
\label{sec:1}

Optical flow, as a motion description of images, has been widely used in high-level video tasks~\cite{xu2019inpaint,yang2019vos,zhu2017deep,cheng2017segflow,chen2017coherent,nilsson2018semantic}.
Benefitting from the growth of deep learning, learning-based optical flow methods~\cite{sun2019pwcpami, neoral2018continual} with considerable accuracy and efficient inference are gradually replacing the classical variational-based approaches~\cite{ren2019mff, maurer2018proflow, wulff2017optical}. However, it is tough to collect the ground truth of dense optical flow in reality, which makes most supervised methods heavily dependent on the large-scale synthetic datasets~\cite{dosovitskiy2015flownetflyingchairs,mayer2016flyingthings}, and the domain difference leads to an underlying degradation when the model is transferred to the real-world.

\begin{figure}[t]
  \centering
  \includegraphics[width=0.85\columnwidth]{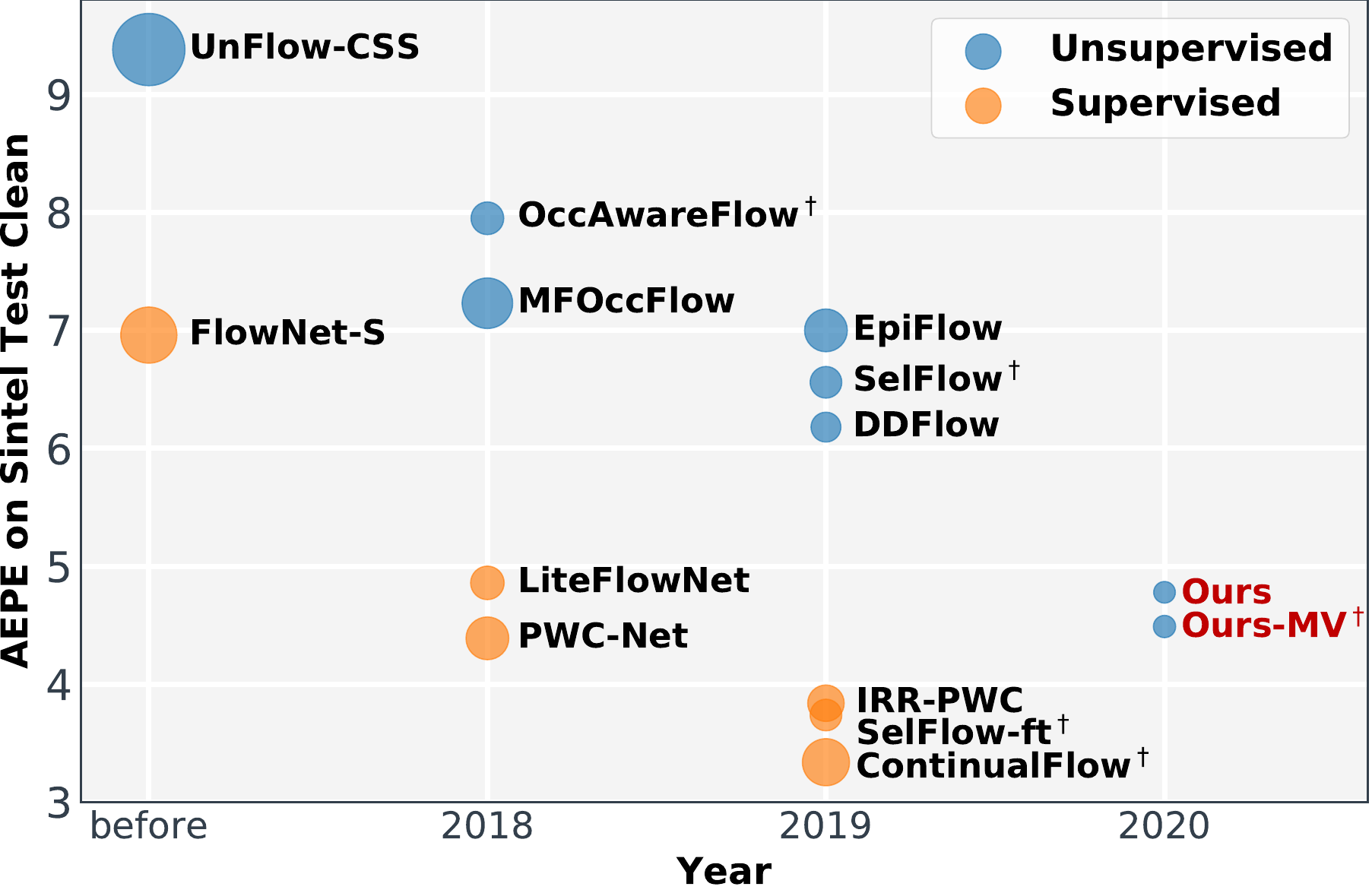}
  \caption{{Timeline of average end-point error (AEPE) advances in deep optical flow.} Marker size indicates network size, and oversized markers have been adjusted. Our method outperforms all of the previous unsupervised methods, also yields comparable accuracy to supervised methods while with fewer parameters. $^{\dagger}$ indicates the model using more than two frames.}
  \label{fig:performance}
  \vspace{-10pt}
\end{figure}

In another point of view, many works proposed to learn optical flow in an unsupervised way~\cite{ren2017unflow17, meister2018unflow,wang2018occaware,liu2019selflow}, in which the ground truth is not necessary. These works aim to train networks with objective from view synthesis~\cite{zhou2017unsupervised,yin2018geonet}, \ie optimizing the difference between reference images and the flow warped target images. This objective is based on the assumption of brightness constancy, which will be violated for challenging scenes, \eg with extreme brightness or partial occlusion. Hence, proper regularization such as occlusion handling~\cite{wang2018occaware,janai2018mfocc} or local smooth~\cite{meister2018unflow} is required. Recent studies have focused on more complicate regularizations such as 3D geometry constraints~\cite{ranjan2019ccflow, wang2019unos, liu2019unsupervised} and global epipolar constraints~\cite{zhong2019epiflow}. As shown in~\cref{fig:performance}, there is still a large gap between these works and supervised methods.
In this paper, we do not rely on the geometrical regularizations but rethink the task itself to improve accuracy.

Interestingly, we notice that almost all of the unsupervised works, such as~\cite{wang2018occaware,liu2019selflow,wang2019unos}, avoid using a heavy combination of augmentations, even if it has been proven effective in supervised flow works~\cite{ilg2017flownet2,sun2018pwc,hur2019irr}.
The reason we conclude is two-fold: \emph{(i)} Data augmentation is essentially a trade-off between diversity and validity. It can improve the model by increasing the diversity of data, while also leads to a shift of data distribution which decreases the accuracy. In unsupervised learning, the benefit of diversity is limited since the abundant training data is easy to access.  \emph{(ii)}
Data augmentation will generate challenging samples, for which view synthesis is more likely to be unreliable, so the objective cannot guide networks for a correct solution.

More recently, there are some works based on knowledge distillation that alleviate the problem of unreliable objective in occluded regions~\cite{liu2019ddflow,liu2019selflow}. The training of these methods is split into two stages. In the first stage, a teacher model is trained to make predictions on original data, and offline creating occluded samples with random crop or mask out. In the second stage, these artificial samples from the teacher model are used to update a student model. However, these methods were designed for the case of partial occluded only. Hence we ask: \emph{Can we generalize the distillation of occlusion to other transformation cases?} Moreover, the distillation method has a bottleneck due to the frozen teacher model. We thus ask: \emph{Can we jointly optimize teacher model and student model, or just training a single network?}

In this work, we address the above two questions with a novel unsupervised learning framework of optical flow. Specifically, for the first question, diverse transformations are used to generate challenging scenes such as low-light, overexposed, with large displacement or partial occlusion. For the second question, instead of optimizing two models with distillation, we simply twist the training step in the regular learning framework by running an additional forward with the input of transformed images, and the transformed flow from the first forward pass is treated as reliable supervision.
Since the self-supervision from transformations avoids the unsupervised objective to be ambiguous in challenging scenes,
our framework allows the network to learn by analogy with the original samples, and gradually mastering the ability to handle challenging samples.

In summary, our contributions are: \emph{(i)} We propose a novel way to make use of the self-supervision signal from abundant augmentations for unsupervised optical flow by only training a single network;
\emph{(ii)} We demonstrate the applicability of our method for various augmentation methods. In addition to occlusion, we develop a general form for more challenging transformations.
\emph{(iii)} Our method leads in a leap of performance among deep unsupervised methods. It also achieves a comparable performance \wrt previous supervised methods, but with much fewer parameters and excellent cross dataset generalization capability.

\section{Related Work}

\paragraph{Supervised Optical Flow.}
Starting from FlowNet~\cite{dosovitskiy2015flownetflyingchairs}, various networks for optical flow with supervised learning have been proposed, \eg FlowNet2~\cite{ilg2017flownet2}, PWC-Net~\cite{sun2018pwc}, IRR-PWC~\cite{hur2019irr}. These methods are comparable in accuracy to well-designed variational methods~\cite{ren2019mff, maurer2018proflow}, and are more effective during inference. However, the success of supervised methods heavily dependent on the large scale synthetic datasets~\cite{mayer2016flyingthings, dosovitskiy2015flownetflyingchairs}, which leads to an underlying degradation when transferring to real-world applications. As an alternative, we dig into the unsupervised method to alleviate the need for ground truth of dense optical flow.

\myparagraph{Unsupervised Optical Flow.} Yu \etal~\cite{jason2016back} first introduced a method for learning optical flow with brightness constancy and motion smoothness, which is similar to the energy minimization in conventional methods. Further researches improve accuracy through occlusion reasoning~\cite{wang2018occaware, meister2018unflow}, multi-frame extension~\cite{janai2018mfocc, guan2019unsupervised}, epipolar constraint~\cite{zhong2019epiflow}, 3D geometrical constraints with monocular depth~\cite{zou2018df, yin2018geonet, ranjan2019ccflow} and stereo depth~\cite{wang2019unos, liu2019unsupervised}. Although these methods have become complicated, there is still a large gap with state-of-the-art supervised methods. Recent works improve the performance by learning the flow of occluded pixels in a knowledge distillation manner~\cite{liu2019ddflow, liu2019selflow}, while the two-stage training in these works is trivial. Instead of studying the complicated geometrical constraints, our approach focuses on the basic training strategy. It generalizes the case of occlusion distillation to more kinds of challenging scenes with a straightforward single-stage learning framework.

\myparagraph{Learning with Augmentation.} Data augmentation is one of the easiest ways to improve training. Recently, there has been something new about integrating augmentation into the learning frameworks. Mounsaveng~\etal~\cite{mounsaveng2019adversarial} and Xiao \etal~\cite{xiao2018spatially} suggested learning data augmentation with a spatial transformer network~\cite{jaderberg2015spatial} to generate more complex samples. Xie~\etal~\cite{xie2019unsupervised} proposed to use augmentation in the semi-supervised tasks by consistency training. Peng~\etal~\cite{peng2018jointly} introduced to optimize data augmentation with the training of task-specific networks jointly. As a new trend in AutoML, several efforts to automatically search for the best policy of augmentations~\cite{cubuk2019autoaugment,ho2019population,lim2019fast} are proposed. All these methods aimed at supervised or semi-supervised learning. In this work, we present a simple yet effective approach to integrate abundant augmentations with unsupervised optical flow. We propose to use reliable predictions of original samples as a self-supervision signal to guide the predictions of augmented samples.

\section{Preliminaries}
\label{sec:3}
\begin{figure}[t]
  \centering
  \begin{overpic}[width=0.99\columnwidth]{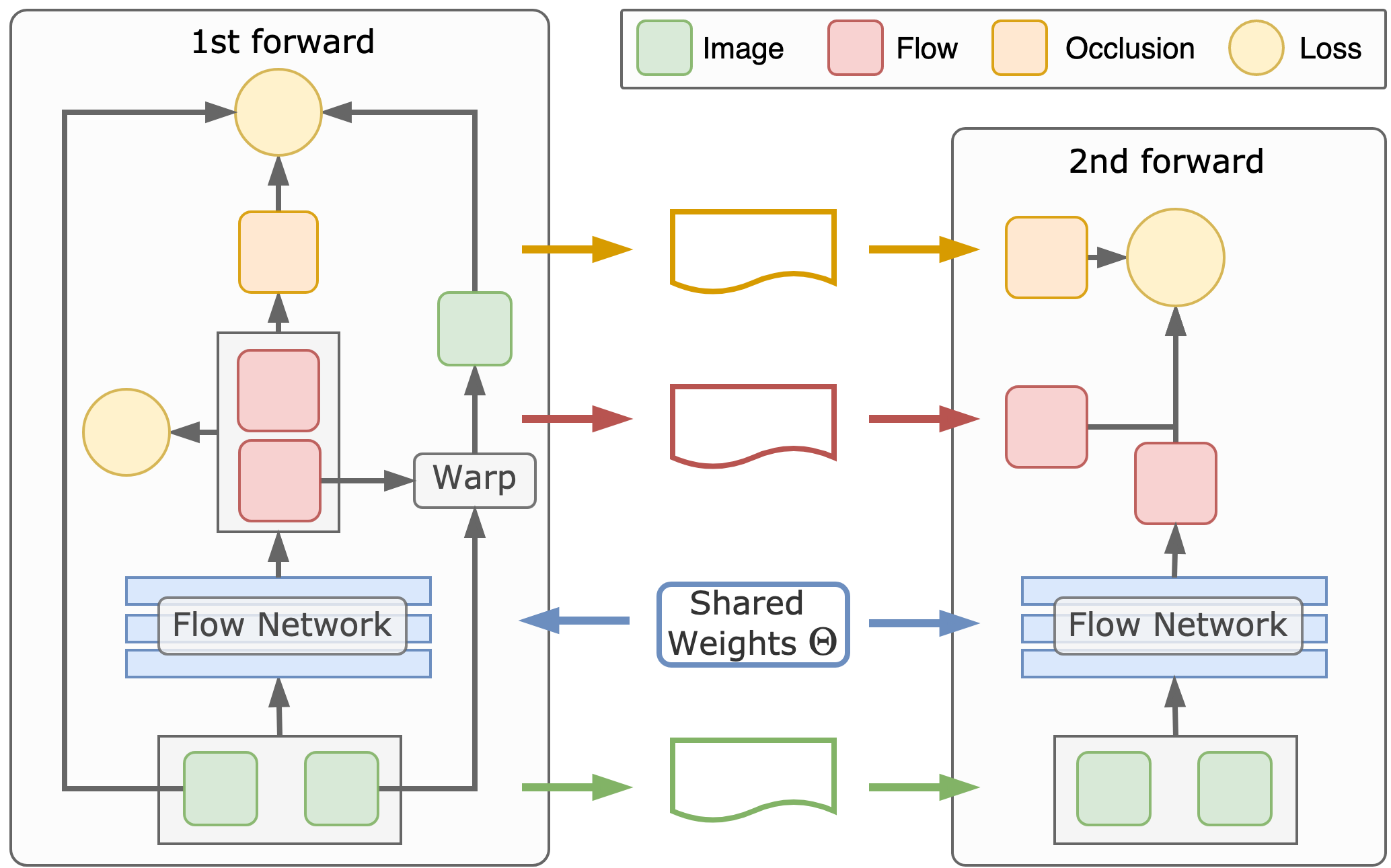}
		\put(14.5, 4.5){\scriptsize $\mathbf{I}_1$}
		\put(23, 4.5){\scriptsize $\mathbf{I}_2$}

		\put(17.4, 26.8){\notsotiny $\mathbf{U}_{12}$}
		\put(17.4, 33.5){\notsotiny $\mathbf{U}_{21}$}
		\put(17.4, 43.5){\notsotiny $\mathbf{O}_{21}$}

		\put(32.6, 37.5){\scriptsize $\hat{\mathbf{I}}_{1}$}

		\put(6.3, 30.4){\footnotesize $\mathcal{L}_\text{sm}$}

		\put(17.5, 53.5){\footnotesize $\mathcal{L}_\text{ph}$}

		\put(51, 5.5){\scriptsize $\mathcal{T}_\theta^\text{img}$}
		\put(50.4, 1.0){\scriptsize \cref{eq:t_img}}
		\put(51, 30.8){\scriptsize $\mathcal{T}_\theta^\text{flo}$}
		\put(50.4, 26.3){\scriptsize \cref{eq:t_flo}}
		\put(51, 43.4){\scriptsize $\mathcal{T}_\theta^\text{occ}$}
		\put(50.4, 38.9){\scriptsize \cref{eq:t_occ}}

		\put(78.7, 4.5){\scriptsize $\overline{\mathbf{I}}_1$}
		\put(87.2, 4.5){\scriptsize $\overline{\mathbf{I}}_2$}
		\put(81.9, 26.8){\notsotiny $\overline{\mathbf{U}}^*_{12}$}
		\put(81.2, 43){\footnotesize $\mathcal{L}_\text{aug}$}

		\put(72.6, 30.6){\notsotiny $\overline{\mathbf{U}}_{12}$}
		\put(72.6, 43.2){\notsotiny $\overline{\mathbf{O}}_{12}$}

  \end{overpic}
	\vspace{.75em}
  \caption{The pipeline of our proposed method. A complete training step includes two forwards: \emph{(i)} The left side shows the first forward with original samples by the regular pipeline introduced in~\cref{sec:3}. Then, we perform transformations on images, predicted flow, and occlusion map respectively to construct an augmented sample. \emph{(ii)} The right side shows an additional forward with the input of transformed images, and the output flow is supervised by the flow prediction of original samples.}
  \label{fig:2}
\end{figure}
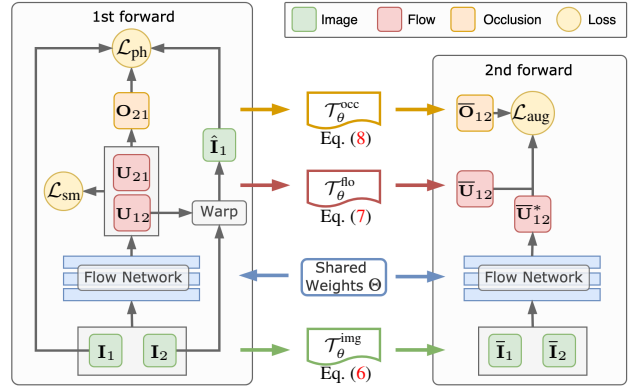

This work aims to learn optical flow from images without the need for ground truth. For completeness, we first briefly introduce the general framework for unsupervised optical flow methods, which is shown in the left part of ~\cref{fig:2}.

Given a dataset of image sequences $\mathcal{I}$, our goal is to train a network $f(.)$ to predict dense optical flow $ \mathbf{U}_{12} $ for two consecutive RGB frames $\left\{\mathbf{I}_{1} , \mathbf{I}_{2}\right\} \in {\mathcal{I}}$,
\begin{equation}
\mathbf{U}_{12} = f(\mathbf{I}_{1}, \mathbf{I}_{2};\Theta),
\end{equation}
where $\Theta$ is the set of learnable parameters in the network.

Despite the lack of direct supervision from ground truth,
the network can be trained implicitly with view synthesis. Specifically, image $ \mathbf{I}_{2} $ can be warped to synthesize the view of $ \mathbf{I}_{1} $ with the prediction of optical flow $\mathbf{U}_{12}$,
\begin{equation}
\hat{\mathbf{I}}_{1}(\mathbf{p})=\mathbf{I}_{2}\left(\mathbf{p}+\mathbf{U}_{12}(\mathbf{p})\right),
\label{eq:2}
\end{equation}
where $\mathbf{p}$ denotes pixel coordinates in the image, and bilinear sampling is used for the continuous coordinates. Then, the objective of view synthesis, also known as photometric loss  $\mathcal{L}_\text{ph}$, can be formulated as:
\begin{equation}
\label{eq:l_ph}
\mathcal{L}_\text{ph} \sim \sum_{\mathbf{p}} \rho(\hat{\mathbf{I}}(\Theta), \mathbf{I}),
\end{equation}
where $\rho(.)$ is a pixel-wise similarity measurement,~\eg $\ell_1$ distance or structural similarities (SSIM).

Nevertheless, the photometric loss is violated when pixels are occluded or moved out of view so that there are no corresponding pixels in $\mathbf{I}_{2}$. As a common practice in~\cite{meister2018unflow,sundaram2010dense}, we denote these pixels by a binary occlusion map $\mathbf{O}_{12}$. This map is obtained by the classical forward-backward checking method, where the backward flow is estimated by swapping the order of input images. The photometric loss in the occluded region will be discarded.

Furthermore, supervision solely based on the photometric loss is ambiguous for somewhere textureless or with repetitive patterns. One of the most common ways to reduce ambiguity is named smooth regularization,
\begin{equation}
\mathcal{L}_\text{sm} \sim \sum_{d \in x, y} \sum_{\mathbf{p}} \left\|\nabla_{d} \mathbf{U}_{12}\right\|_{1} e^{-\left|\nabla_{d} \mathbf{I}\right|},
\label{eq:l_sm}
\end{equation}
which constrains the prediction similar to the neighbors in $x$ and $y$ directions when no significant image gradient exists.

\section{Method}

Since the general pipeline suffers from unreliable supervision for challenging cases, previous unsupervised works avoid using heavy augmentations.
In this section, we introduce a novel framework to reuse existing heavy augmentations that
have been proven effective in the supervised scenario, but with different forms. The pipeline is shown in~\cref{fig:2}, and we will explain in detail next.

\subsection{Augmentation as a Regularization}

Formally, we define an augmentation parameterized by a random vector $\theta$ as $\mathcal{T}_\theta^{\text{img}}: \mathbf{I}_t \mapsto \overline{\mathbf{I}}_t$, from which one can sample augmented images $\{\overline{\mathbf{I}}_1, \overline{\mathbf{I}}_2\}$ based on original images $\{\mathbf{I}_1, \mathbf{I}_2\}$ in the dataset. In the general pipeline, the network is trained with the data sampled from the augmented dataset. In contrast, we train the network on original data, but leverage augmented samples as a regularization.

More specifically, after a regular forward pass for original images, we additionally run another forward for transformed images to predict the optical flow $\overline{\mathbf{U}}^*_{12}$. Meanwhile, the prediction of optical flow in the first forward is transformed consistently by $\mathcal{T}_\theta^{\text{flo}}: \mathbf{U}_{12} \mapsto \overline{\mathbf{U}}_{12}$.

The basic assumption of our method is that augmentation brings challenging scenes in which the unsupervised loss will be unreliable, while the transformed predictions of original data can provide reliable self-supervision. Therefore, we optimize the consistency for the transformed samples instead of the objective of view synthesis. We follow the generalized Charbonnier function that commonly used in the supervised learning of optical flow as:
\begin{equation}
\mathcal{L}_\text{aug} ~\sim \sum_{\mathbf{p}}\left(\left|\mathcal{S}\left(\overline{\mathbf{U}}_{12}(\mathbf{p})\right)-\overline{\mathbf{U}}^*_{12}(\mathbf{p})\right|+\epsilon\right)^{q},
\label{eq:l_aug}
\end{equation}
where $\mathcal{S}(.)$ stands for stop-gradient, and the same setting as supervised work~\cite{sun2018pwc} with $q=0.4$ and $\epsilon=0.01$ gives less penalty to outliers. For stability, we stop the gradients of $\mathcal{L}_\text{aug}$ propagating to the transformed original flow $\overline{\mathbf{U}}_{12}$. Also, only the loss in the non-occluded region is considered. After twice forwarding, the photometric loss~\cref{eq:l_ph}, the smooth regularization~\cref{eq:l_sm}, and the augmentation regularization ~\cref{eq:l_aug} are backward at once to update the model.

Our learning framework can be integrated with almost all types of augmentation methods. In the following, we summarize three kinds of transformations, which compose the common augmentations for the optical flow task. Some examples are shown in~\cref{fig:transformations}.

\newcommand{\imgheight}{0.066\textwidth}
\newcommand{\imgwidth}{0.15\textwidth}
\newcommand{\Csize}{75pt}
\begin{figure}[t]
	\centering
	\scriptsize
	\setlength\tabcolsep{0.5pt}
	\begin{tabular}{@{}lC{\Csize}C{\Csize}C{\Csize}@{}}
  & Target Image $\mathbf{I}_2$ & Predicted Flow $\mathbf{U}_{12}$ & Ground Truth (Unused)\\
  &
  \raisebox{-.5\height}{\includegraphics[height=\imgheight, width=\imgwidth]{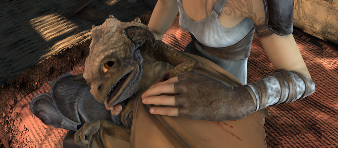}} &
  \raisebox{-.5\height}{\includegraphics[height=\imgheight, width=\imgwidth]{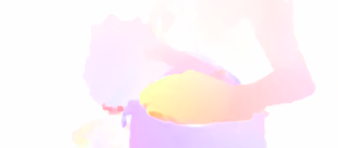}} &
  \raisebox{-.5\height}{\includegraphics[height=\imgheight, width=\imgwidth]{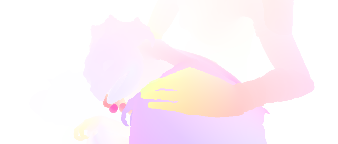}} \\ \midrule
  & Target Image $\overline{\mathbf{I}}_2$ & Predicted Flow $\overline{\mathbf{U}}^*_{12}$ & Transformed Flow $\overline{\mathbf{U}}_{12}$\\
  (a) &
  \raisebox{-.5\height}{\includegraphics[height=\imgheight, width=\imgwidth]{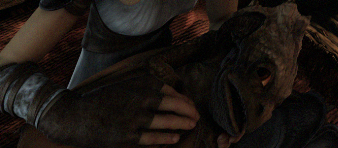}} &
  \raisebox{-.5\height}{\includegraphics[height=\imgheight, width=\imgwidth]{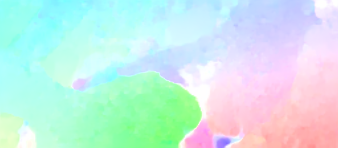}} &
  \raisebox{-.5\height}{\includegraphics[height=\imgheight, width=\imgwidth]{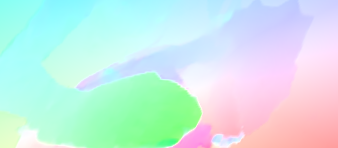}} \\
  \vspace{3pt}
  (b) &
  \raisebox{-.5\height}{\includegraphics[height=\imgheight, width=\imgwidth]{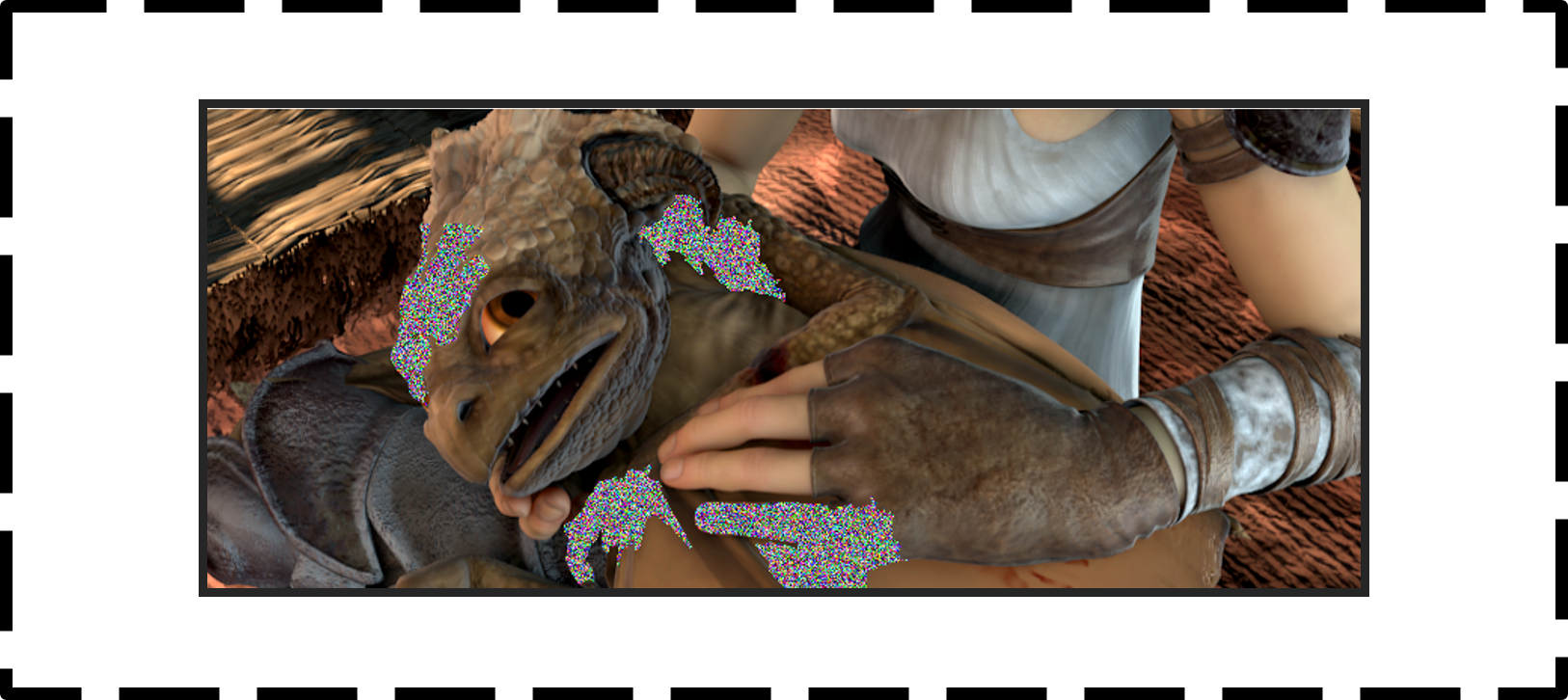}} &
  \raisebox{-.5\height}{\includegraphics[height=\imgheight, width=\imgwidth]{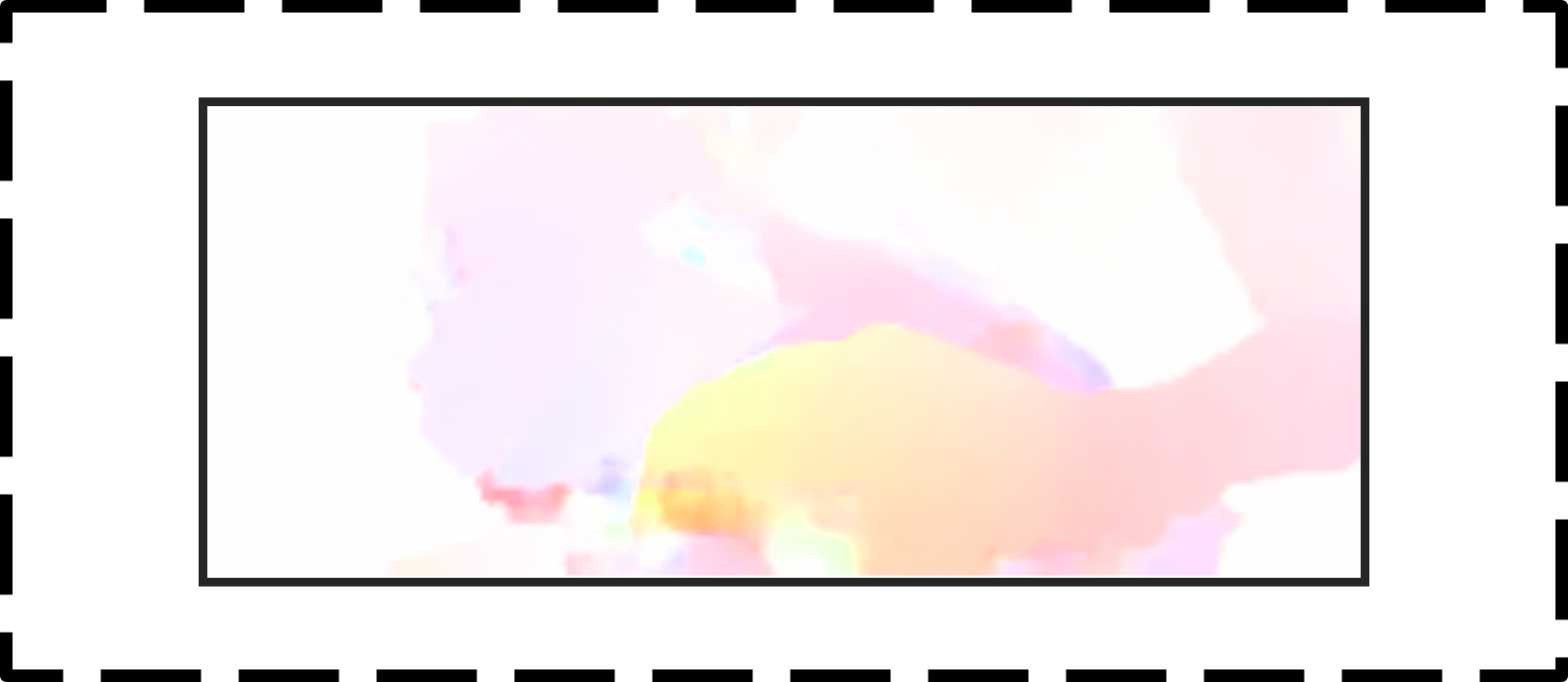}} &
  \raisebox{-.5\height}{\includegraphics[height=\imgheight, width=\imgwidth]{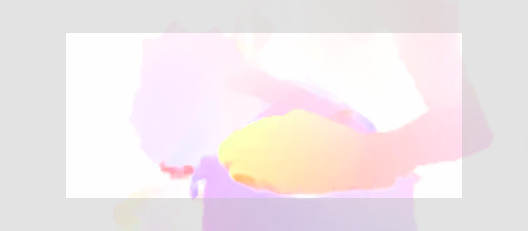}}
  \end{tabular}
  \caption{Some examples of the main idea. The same network is used to predict the optical flow of original images and transformed images, respectively. (a) Spatial transformation and appearance transformation generate a scene with large displacement and low brightness. (b) Occlusion transformation introduces additional occlusions. The pseudo label $\overline{\mathbf{U}}_{12}$ that transformed from the original predictions $\mathbf{U}_{12}$ can provide reliable supervision.}
  \label{fig:transformations}
\end{figure}

\paragraph{Spatial Transformation.}

We assume the transformation that results in a change in the location of pixels is called spatial transformation, which includes random crop, flip, zoom, affine transform, or more complicated transformations such as thin-plate-spline or CPAB transformations~\cite{freifeld2017cpa}.

Here we show a general form for these transformations. Let $\tau_\theta$ be a transformation of pixel coordinates. The transformation of image $\mathcal{T}_\theta^\text{img}: \mathbf{I}_t \mapsto \overline{\mathbf{I}}_t$ can be formulated as:

\begin{equation}
  \overline{\mathbf{I}}_t(\mathbf{p}) = \mathbf{I}_t \left( \tau_\theta(\mathbf{p}) \right),
	\label{eq:t_img}
\end{equation}
which can be implemented by a differentiable warping process, same as the one used in~\cref{eq:2}.

Since changing pixel locations will lead to a change in optical flow, we should warp on an intermediate flow field $\widetilde{\mathbf{U}}_{12}$ instead of the original flow. The transformation of optical flow is $\mathcal{T}_\theta^\text{flo}: \mathbf{U}_{12} \mapsto \overline{\mathbf{U}}_{12}$ can be formulated as:
\begin{equation}
  \left\{
    \begin{array}{l}
    \widetilde{\mathbf{U}}_{12}(\mathbf{p}) = \tau_\theta \left( \mathbf{p} + \mathbf{U}_{12}(\mathbf{p}) \right) -  \tau_\theta \left( \mathbf{p} \right) \vspace{1ex},\\
    \overline{\mathbf{U}}_{12}(\mathbf{p}) = \widetilde{\mathbf{U}} \left( \tau^{-1}_\theta\left(\mathbf{p}\right) \right).
    \end{array}
  \right.
	\label{eq:t_flo}
\end{equation}

 Additionally, the spatial transformation brings new occlusions. As we mentioned above, we explicitly reasoning occlusion from the predictions of bi-directional optical flow. Since predictions of transformed samples are noisy, we infer the transformed occlusion map from original predictions instead. The transformation $\mathcal{T}_\theta^\text{occ}: \mathbf{O}_{12} \mapsto \overline{\mathbf{O}}_{12}$ consists of two parts: the old occlusion $\overline{\mathbf{O}}^\text{old}_{12}(\mathbf{p})$ in the new view and the new occlusion $\overline{\mathbf{O}}^\text{new}_{12}(\mathbf{p})$ for pixels whose correspondences are out of the boundary $\Omega$. The former can be obtained by the same warping process as $\mathcal{T}_{\theta}^\text{img}$ but with nearest-neighbor interpolation, and the latter can be explicitly estimated from the flow $\overline{\mathbf{U}}_{12}$ by checking the boundary:
 \begin{equation}
  \overline{\mathbf{O}}^\text{new}_{12}(\mathbf{p}) = \left(\mathbf{p} + \overline{\mathbf{U}}_{12}(\mathbf{p})\right) \notin \Omega.
  \label{eq:t_occ}
 \end{equation}

The final transformed occlusion $\overline{\mathbf{O}}_{12}$ is a union of these two parts. Note that, the non-occluded pixels in $\overline{\mathbf{O}}^\text{old}_{12}$ might be occluded in $\overline{\mathbf{O}}^\text{new}_{12}$. It provides an effective way to learn the optical flow in occluded regions. For stability, only the non-occluded pixels in $\overline{\mathbf{O}}^\text{old}_{12}$ contribute to the loss $\mathcal{L}_\text{aug}$.

Besides, since we formulate the spatial transformation as a warping process, there might be pixels out of boundary after transformation. The common solution, such as padding with zero or the value of boundary pixels, will lead to severe artifacts. Therefore, we repeat sampling the transformations until all transformed pixels are in the region of the original view. On the other hand, this strategy increases the displacement of the pixel in general.

\begin{figure*}[t]
 \centering
	\begin{overpic}[width=0.95\textwidth]{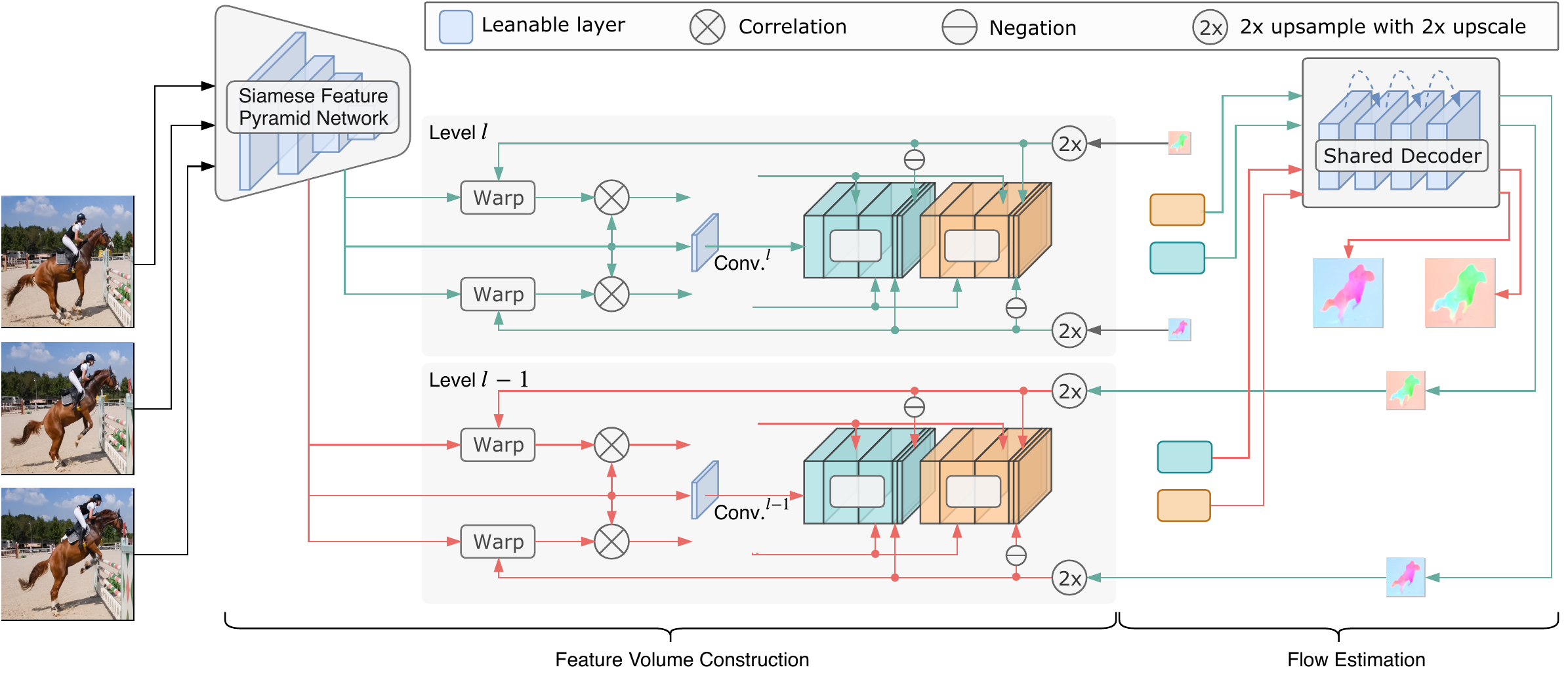}
		\put(9.0, 6.4){\scriptsize $\mathbf{I}_2$}
		\put(9.0, 15.75){\scriptsize $\mathbf{I}_1$}
		\put(9.0, 25.1){\scriptsize $\mathbf{I}_0$}

	 \put(20.4, 15.8){\scriptsize $\mathbf{x}^{l-1}_0$}
	 \put(20.4, 12.6){\scriptsize $\mathbf{x}^{l-1}_1$}
	 \put(20.4, 9.6){\scriptsize $\mathbf{x}^{l-1}_2$}

	 \put(22.4, 31.5){\scriptsize $\mathbf{x}^{l}_0$}
	 \put(22.4, 28.3){\scriptsize $\mathbf{x}^{l}_1$}
	 \put(22.4, 25.3){\scriptsize $\mathbf{x}^{l}_2$}

	 \put(44.4, 14.8){\scriptsize $\mathbf{cv}^{l-1}_{10}$}
	 \put(44.4, 8.2){\scriptsize $\mathbf{cv}^{l-1}_{12}$}
	 \put(44.4, 30.6){\scriptsize $\mathbf{cv}^{l}_{10}$}
	 \put(44.4, 24){\scriptsize $\mathbf{cv}^{l}_{12}$}

	 \put(91.2, 16.8){\scriptsize $\mathbf{U}^{l}_{10}$}
	 \put(91.2, 7.4){\scriptsize $\mathbf{U}^{l}_{12}$}

	 \put(71., 32.2){\scriptsize $\mathbf{U}^{l+1}_{10}$}
	 \put(71., 23.2){\scriptsize $\mathbf{U}^{l+1}_{12}$}

	 \put(84.3, 20.7){\scriptsize $\mathbf{U}^{l-1}_{12}$}
	 \put(91., 20.7){\scriptsize $\mathbf{U}^{l-1}_{10}$}

   \put(52.95, 11.75){\tiny $F^{l-1}_{10}$}
   \put(60.4, 11.75){\tiny $F^{l-1}_{12}$}

   \put(53.3, 27.45){\tiny $F^{l}_{10}$}
   \put(60.75, 27.4){\tiny $F^{l}_{12}$}

   \put(73.8, 29.7){\tiny $F^{l}_{12}$}
   \put(73.8, 26.65){\tiny $F^{l}_{10}$}

   \put(73.8, 13.95){\tiny $F^{l-1}_{10}$}
   \put(73.8, 10.85){\tiny $F^{l-1}_{12}$}

 \end{overpic}
 \caption{Network architecture for our lightweight multi-frame extension of PWC-Net~\cite{sun2018pwc}. It shares a semi-dense flow decoder for all of the levels across the pyramid with both forward flow and backward flow. For simplicity and completeness, the pipeline of two levels in the feature pyramid is displayed. Different line colors represent different levels of the process.}
 \label{fig:network}
\end{figure*}
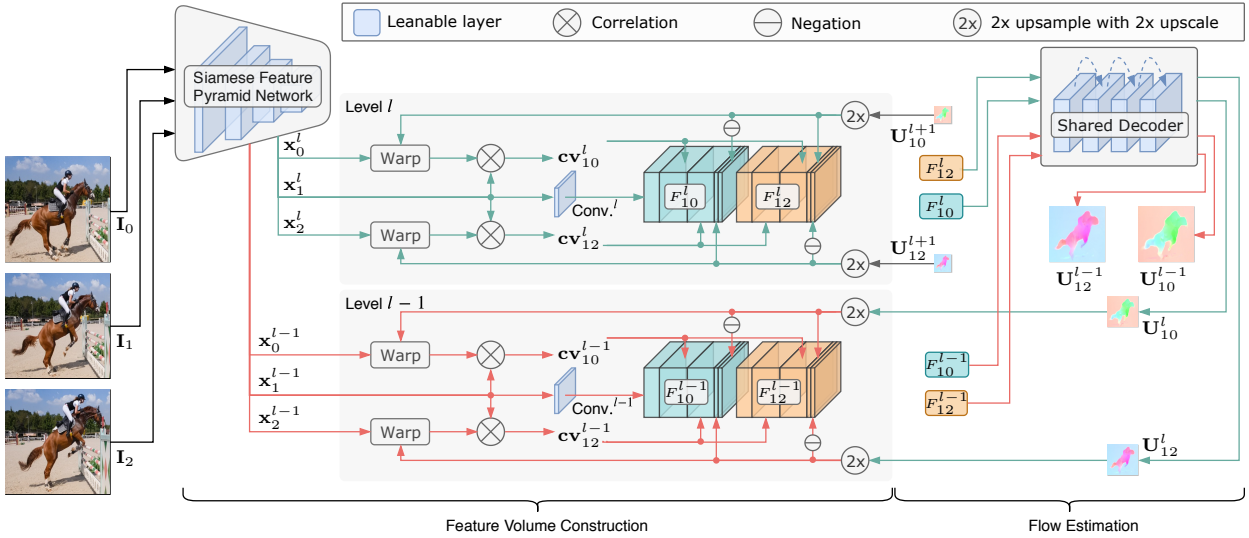

\vspace{.5em}
\myparagraph{Occlusion Transformation.}
The spatial transformation provides reliable supervision for the flow with large displacement or occlusion around the boundary. As a complementary, recent work~\cite{liu2019ddflow, liu2019selflow} proposed to learn optical flow in arbitrary occluded regions with knowledge distillation. The general learning process of these methods consists of training a teacher model, offline creating occluded samples, and distilling to a student model. We argue that the way of model distillation is too trivial, and there is a performance bottleneck due to the frozen teacher model.

We integrate the occlusion hallucination into our one-stage training framework and named as occlusion transformation. Specifically, there are two steps: \emph{(i)} Random crop. Actually, random crop is a kind of spatial transformation, but it efficiently creates new occlusion in the boundary. We crop the pair of images as a preprocess of occlusion transformation. \emph{(ii)} Random mask out. We randomly mask out some superpixels in the target images with Gaussian noise, which will introduce new occlusion for the source image.

Note that, we adopt a strategy consistent with the spatial transformation that only the pixels not occluded in $\overline{\mathbf{O}}^\text{old}_{12}$ contribute to $\mathcal{L}_\text{aug}$. It is different from the previous distillation works, in which they reasoning a new occlusion map from the noisy prediction of transform images. Besides, in order to avoid creating transformed samples offline, we adopt a fast method of superpixel segmentation similar to~\cite{gSLICr_2015}. The occlusion transformation in our framework simplifies the way of model distillation by optimizing a single model in one-stage with end-to-end learning.

\vspace{.5em}
\myparagraph{Appearance Transformation.}
More transformations only change the appearance of images, such as random color jitter, random brightness, random blur, random noise. As a relatively simple case, appearance transformation does not change the location of pixels, nor introduce new occlusion. Still, the transformations lead to a risk for general methods, \eg the photometric loss is meaningless when the image is overexposed, blurred, or in extremely low light. Instead, our method can exploit these transformations since the prediction of the original sample provides a way to learn the optical flow in challenging transformed scenes.

\subsection{Overall Objective and Convergence Analysis}

Our framework assumes that transformed predictions are generally more accurate than the predictions of transformed samples, but what if samples are in the opposite case? In fact, we ensure convergence with the scope of each loss, \ie, which pixels affect each loss.

As shown in~\cref{fig:2}, the overall objective for a training step consists of three loss terms in twice forwarding,
\begin{equation}
\resizebox{0.9\hsize}{!}{
$ \mathcal{L}_{\text {all }}=\underbrace{\mathcal{L}_{\mathrm{ph}}\left(\mathbf{U}_{12}\right)+\lambda_{1} \mathcal{L}_{\mathrm{sm}}\left(\mathbf{U}_{12}\right)}_{\text {1st forward }}+\underbrace{\lambda_{2} \mathcal{L}_{\text {aug}}(\mathcal{S}(\overline{\mathbf{U}}_{12}), \overline{\mathbf{U}}_{12}^{*})}_{\text {2nd forward}},$
}
\end{equation}
in which the first two terms propagate gradients for the original sample, and the last term is for the transformed sample.
The original data and the augmented data are treated differently. By setting a minor weight of $\lambda_2$, we can ensure that the original data is always dominant, so the effects of bad cases are limited. Moreover, the scope of the photometric loss $\mathcal{L}_\text{ph}$ is the non-occluded pixels in $\mathbf{O}_{12}$. Thus the augmentation consistency loss becomes dominant for the new occluded pixels, which leads the network to learn the optical flow with occlusion effectively. Besides, the scope of augmentation loss $\mathcal{L}_\text{aug}$ avoids the network to be misguided from the original occluded predictions.

\subsection{Lightweight Network Architecture}
The learning framework we proposed can be applied to any flow networks. However, optical flow often plays a role as a sub-module in high-level video tasks~\cite{xu2019inpaint,yang2019vos,nilsson2018semantic} where the model size should be concerned.
Hence, we introduce a lightweight architecture and extend it to multiple frames.

We start from a well-known network for the optical flow task named PWC-Net~\cite{sun2018pwc}. The original network shares a feature encoder with a siamese feature pyramid network for the images. For the level $l$ in the pyramid, the feature maps of target image $\mathbf{x}_2^l$ are aligned by warping operation with the flow prediction ${\mathbf{U}}_{12}^{l+1}$ from the higher level. Then the cost volume $\mathbf{cv}_{12}^{l}$ is constructed with correlation operation. The input for flow decoder $F_{12}^l$ is organized by concatenating the feature maps of source image $\mathbf{x}_1^l$, the upsampled flow from the higher level $\mathbf{U}_{12}^{l+1}$, and the cost volume $\mathbf{cv}^l_{12}$. Finally, the specific flow decoder of level $l$ predicts the optical flow $\mathbf{U}_{12}^l$. By iterating over the pyramid, the network predicts optical flow at different scales.

Our method follows the main pipeline of the original PWC-Net but with some modifications. The flowchart of our multi-frame extension is shown in~\cref{fig:network}.
We notice that the majority of learnable parameters of PWC-Net is in the flow decoder of each feature level, so we take several steps to reduce the parameters: \emph{(i)} The original implementation adopts a fully dense connection in each decoder, while we reduce the connections that only connections in the nearest two layers are retained. \emph{(ii)} We share the flow decoder for all of the levels across the pyramid, with an additional convolution layer for each level to align the feature maps. \emph{(iii)} We extend the model to multiple frames by repeating the warping and correlation to the backward features. The flow decoder is shared for both forward flow and backward flow in the multi-frame extension by changing the sign of optical flow and the order in feature concatenation.

\section{Experimental Results}
\subsection{Implementation Details}
We implement our end-to-end approach in PyTorch~\cite{paszke2017pytorch}. All models are trained by Adam optimizer~\cite{kingma2014adam} with $\beta_1=0.9$, $\beta_2=0.99$, batch size of 4. The learning rate is $10^{-4}$ without adjustment during training.
The loss weights for regularizations are set to $\lambda_1 = 60$ and $\lambda_2=0.01$ for all datasets.
In addition, an optional pre-training can be used for better results,
which is under almost the same setting above, but with $\lambda_2=0$, \ie a regular training step without the transformed pass in forward~\footnote{Code available at \url{https://github.com/lliuz/ARFlow}.}.

Only random flip and random time order switch are performed as the regular data augmentation. The heavy combination of augmentations in supervised works~\cite{ilg2017flownet2,sun2018pwc,Hui2018liteflownet} are used as the appearance transformation and spatial transformation in our framework, including random rotate, translate, zoom in, as well as additive Gaussian noise, Gaussian blur and random jitter in brightness, color, and contrast.

\subsection{Datasets}
We first evaluate our method on three well-established optical flow benchmarks, MPI Sintel~\cite{butler2012sintel}, KITTI 2012~\cite{Geiger2012kitti12}, and KITTI 2015~\cite{Menze2015kitti15}. Then, we conduct a cross dataset experiment with another optical flow dataset FlyingChairs~\cite{dosovitskiy2015flownetflyingchairs} and a segmentation dataset CityScapes~\cite{Cordts2016Cityscapes}.

We follow a similar data setting in previous unsupervised works~\cite{liu2019ddflow, liu2019selflow}. For the MPI Sintel benchmark, we extract all frames from the raw movie and manually group frames by shots for pre-training, which consists of 14,570 image pairs. Then, the model is fine-tuned on the standard training set, which provides 1,041 image pairs with two different rendering passes (``Clean'' and ``Final''). For the KITTI 2012 and KITTI 2015, we pre-train the model on the KITTI raw dataset~\cite{Geiger2013kittiraw}, but discard scenes that contain images appeared in the optical flow benchmarks. The pre-training set consists of 28,058 image pairs. Then the model is fine-tuned on the multi-view extension data, but discards samples containing frames related to validation, \ie numbers 9-12. The final training set consists of 6,000 samples for our basic model and 3,600 samples for the multi-frame model.

\subsection{Comparison with State-of-the-art}

\begin{table}[t]
\centering
\scriptsize
\begin{tabular*}{\columnwidth}{@{}p{1pt}lccccc@{}}
\toprule
& \multicolumn{1}{l}{\multirow{2}{*}{Method}} & \multicolumn{2}{c}{Sintel Training} & \multicolumn{2}{c}{Sintel Test} & \multicolumn{1}{l}{\multirow{2}{*}{{\# Param.}}} \\
\cmidrule(lr){3-4} \cmidrule(lr){5-6}
& \multicolumn{1}{c}{}                        & Clean            & Final            & Clean          & Final          \\ \midrule
\multirow{5}{*}{\rotatebox[origin=c]{90}{Supervised}}
& FlowNetS-ft~\cite{dosovitskiy2015flownetflyingchairs}                                 & (3.66)           & (4.44)           & 6.96           & 7.76     & 32.07 M      \\
& LiteFlowNet-ft\cite{Hui2018liteflownet}                              & \textbf{(1.64)}           & (2.23)           & 4.86           & 6.09     & 5.37 M      \\
& PWC-Net-ft\cite{sun2018pwc}                                  & (2.02)           & (2.08)           & 4.39           & 5.04     & 8.75 M      \\
& IRR-PWC-ft~\cite{hur2019irr}                              & (1.92)           & (2.51)           & 3.84           & 4.58     & 6.36 M \\
& SelFlow-ft$^{\dagger}$~\cite{liu2019selflow}                   & (1.68)           & \textbf{(1.77)}           & \textbf{3.74}           & \textbf{4.26}  & \textbf{4.79 M} \\
\midrule
\multirow{8}{*}{\rotatebox[origin=c]{90}{Unsupervised}}
& UnFlow-CSS~\cite{meister2018unflow}                               & -                & (7.91)           & 9.38           & 10.22        &  116.58 M  \\
& OccAwareFlow~\cite{wang2018occaware}                             & (4.03)           & (5.95)           & 7.95           & 9.15    & 5.12 M       \\
& MFOccFlow$^{\dagger}$~\cite{janai2018mfocc}                   & (3.89)           & (5.52)           & 7.23           & 8.81    & 12.21 M       \\
& EpiFlow train-ft~\cite{zhong2019epiflow}                         & (3.54)           & (4.99)           & 7.00           & 8.51      & 8.75 M     \\
& DDFlow~\cite{liu2019ddflow}                                   & (2.92)           & (3.98)           & 6.18           & 7.40      & 4.27 M     \\
& SelFlow$^{\dagger}$~\cite{liu2019selflow}                      & (2.88)           & (3.87)           & 6.56           & 6.57      & 4.79 M     \\
& \bfseries Ours (ARFlow)                                    & (2.79)  & (3.73)  & 4.78  & 5.89  & \textbf{2.24 M} \\
& \bfseries Ours (ARFlow-MV$^{\dagger}$)                      & \textbf{(2.73)}           & \textbf{(3.69)}           & \textbf{4.49}        & \textbf{5.67} & 2.37 M  \\ \bottomrule
\end{tabular*}
\caption{\textbf{MPI Sintel Flow}: AEPE and the number of CNN parameters are reported.
Missing entry (-) means that the results are not reported for the respective method, and $^{\dagger}$ indicates the model using more than two frames.}
\label{tab:flow_Sintel}
\vspace{-.5em}
\end{table}

\begin{table}[t]
\centering
\scriptsize
\begin{tabularx}{\columnwidth}{@{}p{1pt}XcccS[table-format=2.2,table-space-text-post = \si{\%}]@{}}
	\toprule
	& \multirow{3}{*}{Method}  &  \multicolumn{2}{c}{KITTI 2012} & \multicolumn{2}{c}{KITTI 2015}  \\ \cmidrule(lr){3-4} \cmidrule(l){5-6}
  &    					    			 	 & training 	& test               & training & \text{test (F1)} \\ \midrule
	\multirow{4}{*}{\rotatebox[origin=c]{90}{Supervised}}
  & FlowNet2-ft~\cite{ilg2017flownet2}			 		      & (1.28) & 1.8 & (2.30) & 11.48\% \\
  & LiteFlowNet-ft~\cite{Hui2018liteflownet}		& (1.26) & 1.7 & (2.16) & 11.48 \% \\
	& PWC-Net-ft~\cite{sun2018pwc}				& (1.45) & 1.7 & (2.16) & 9.60 \% \\
	& SelFlow-ft$^{\dagger}$~\cite{liu2019selflow}    & \textbf{(0.76)} & \textbf{1.5} & \textbf{(1.18)} & \bfseries8.42 \% \\   \midrule
 \multirow{8}{*}{\rotatebox[origin=c]{90}{Unsupervised}}
  & BridgeDepthFlow$^\S$~\cite{lai2019bridging}      & 2.56   & --  & 7.02 & {--}\\
	& CCFlow$^\S$~\cite{ranjan2019ccflow}               & --     & --  & 5.66 & 25.27\%\\
	& UnOS-stereo$^{\S}$~\cite{wang2019unos}			  & 1.64   & 1.8 & 5.58 & 18.00\% \\
	& EpiFlow-train-ft$^\S$~\cite{zhong2019epiflow}			& (2.51) & 3.4 &(5.55)& 16.95\% \\

  & DDFlow~\cite{liu2019ddflow}			  & 2.35   & 3.0 & 5.72 & 14.29\% \\
	& SelFlow$^{\dagger}$~\cite{liu2019selflow}			  & 1.69   & 2.2 & 4.84& 14.19\% \\
  & \bfseries Ours (ARFlow)                     & 1.44  & 1.8          & \textbf{2.85} & 11.80\% \\
  & \bfseries Ours (ARFlow-MV$^{\dagger}$)      & \textbf{1.26}  & \textbf{1.5} & 3.46 & \bfseries 11.79\% \\ \bottomrule
\end{tabularx}
\caption{\textbf{KITTI Optical Flow 2012 and 2015}: AEPE and Fl are reported. For unsupervised methods, only the works published in 2019 are shown. Missing entry (-) means that the results are not reported for the respective method. $^{\dagger}$ indicates the model using more than two frames. $\S$ indicates training with geometrical constraints.}
\label{tab:flow_KITTI}
\vspace{-1em}
\end{table}

We compare our method with both supervised and unsupervised methods on optical flow benchmarks.
Standard metrics for optical flow are used, including average end-point error (AEPE), and percentage of erroneous pixels (Fl).

~\cref{tab:flow_Sintel} reports the results on MPI Sintel benchmark. Our basic two-frame model ``ARFlow'' outperforms all previous unsupervised works with the least parameters. Furthermore, our multi-frame model ``ARFlow-MV'' reduces the previous best AEPE from 6.18~\cite{liu2019ddflow} to 4.49 on the clean pass, with 27.3\% improvement, and from 6.57~\cite{liu2019selflow} to 5.67 on the final pass, with 13.7\% improvement.

As for KITTI benchmarks, ~\cref{tab:flow_KITTI} shows a significant improvement. On the training set, we achieve AEPE=1.26 with 25.4\% relative improvement on KITTI 2012 and AEPE=2.85 with 41.2\% improvement on KITTI 2015~\wrt the previous best unsupervised method~\cite{liu2019selflow}. On the test set, our method reaches the best AEPE=1.5 and F1-all=11.79\% among unsupervised methods, respectively.

Several representative supervised methods are also reported as a reference. As a result, our unsupervised models firstly reach or approach some powerful fully supervised methods such as LiteFlowNet~\cite{Hui2018liteflownet}, PWC-Net~\cite{sun2018pwc}, even with 27.1\% parameters of PWC-Net.

\newcommand{\sintelheight}{0.073\textwidth}
\newcommand{\kittiheight}{0.058\textwidth}
\newcommand{\viswidth}{0.19\textwidth}
\newcommand{\visCsize}{95pt}

\begin{figure*}[t]
	\centering
	\scriptsize
	\setlength\tabcolsep{0.5pt}
	\begin{tabular}{@{}C{\visCsize}C{\visCsize}C{\visCsize}C{\visCsize}C{\visCsize}@{}}
		(a) Reference Image & (b) Our Predictions & (c) SelFlow~\cite{liu2019selflow} Predictions& (d) Our Error & (e) SelFlow Error~\cite{liu2019selflow} \\
		\includegraphics[height=\sintelheight, width=\viswidth]{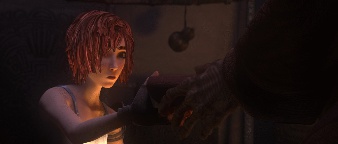} &
		\includegraphics[height=\sintelheight, width=\viswidth]{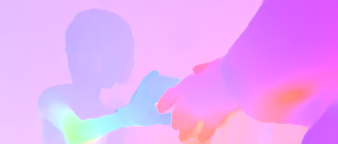} &
		\includegraphics[height=\sintelheight, width=\viswidth]{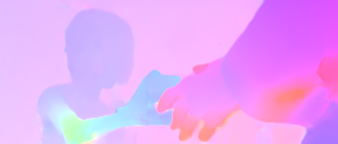} &
		\includegraphics[height=\sintelheight, width=\viswidth]{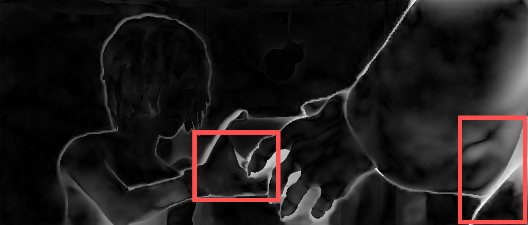} &
		\includegraphics[height=\sintelheight, width=\viswidth]{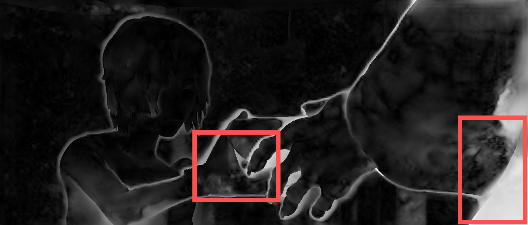}
		\\
		\includegraphics[height=\sintelheight, width=\viswidth]{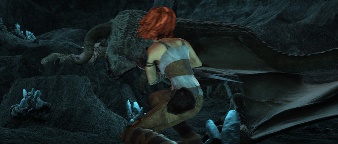} &
		\includegraphics[height=\sintelheight, width=\viswidth]{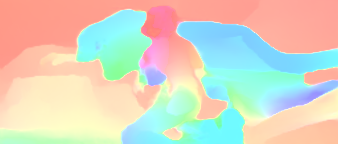} &
		\includegraphics[height=\sintelheight, width=\viswidth]{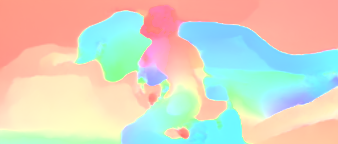} &
		\includegraphics[height=\sintelheight, width=\viswidth]{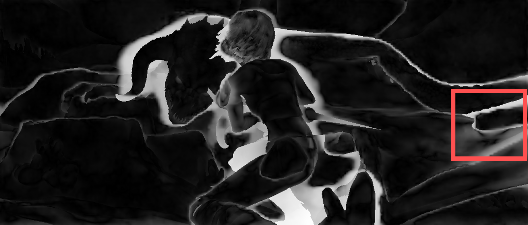} &
		\includegraphics[height=\sintelheight, width=\viswidth]{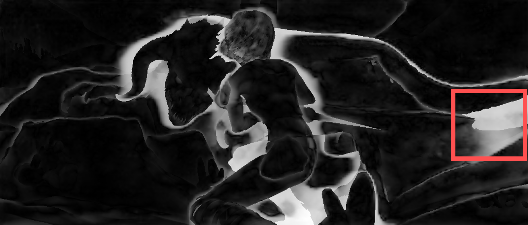}
		\\
		\includegraphics[height=\kittiheight, width=\viswidth]{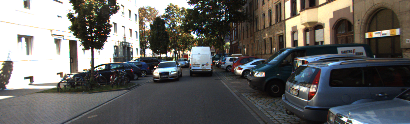} &
		\includegraphics[height=\kittiheight, width=\viswidth]{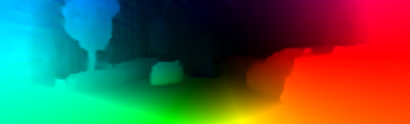} &
		\includegraphics[height=\kittiheight, width=\viswidth]{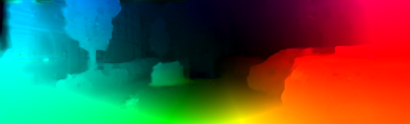} &
		\includegraphics[height=\kittiheight, width=\viswidth]{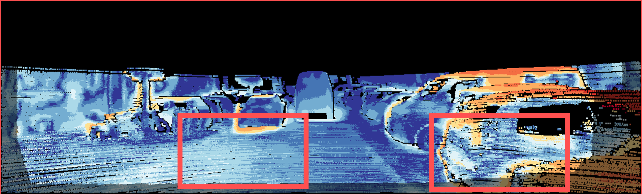} &
		\includegraphics[height=\kittiheight, width=\viswidth]{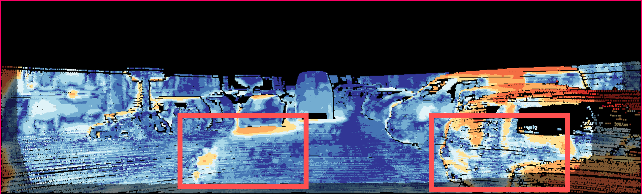}
		\\
		\includegraphics[height=\kittiheight, width=\viswidth]{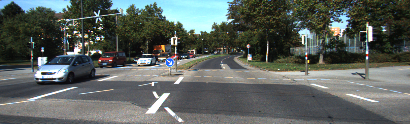} &
		\includegraphics[height=\kittiheight, width=\viswidth]{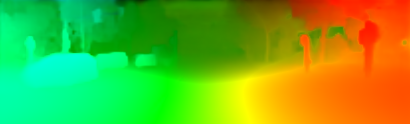} &
		\includegraphics[height=\kittiheight, width=\viswidth]{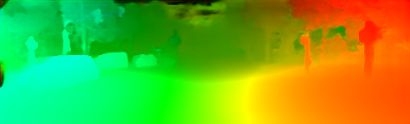} &
		\includegraphics[height=\kittiheight, width=\viswidth]{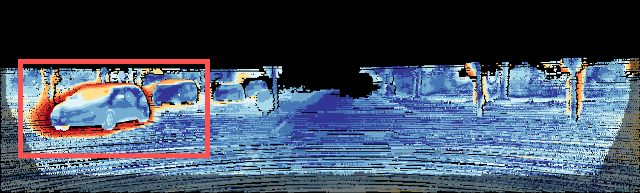} &
		\includegraphics[height=\kittiheight, width=\viswidth]{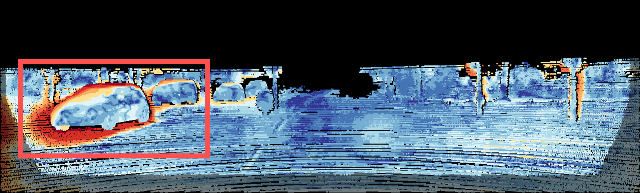}
	\end{tabular}

  \vspace{-3pt}
	\caption{Qualitative visualization comparing with unsupervised SelFlow~\cite{liu2019selflow}. The first two rows are from the Sintel Final pass, where the errors are visualized in gray. The last two rows are from KITTI 2015, in which the correct predictions are depicted in blue and the wrongs in red for the error visualization. More samples will be available on the website of corresponding benchmarks.
	}
	\vspace{-10pt}
	\label{fig:visualization}
\end{figure*}

Samples on MPI Sintel and KITTI are shown in~\cref{fig:visualization}. Compared with the state-of-the-art competitor~\cite{liu2019selflow}, for the low light and large displacement scenes in MPI Sintel, our method maintains better performance in general and is more accurate around the boundaries. For KITTI results, the shapes in our optical flow are more structured for objects and more accurate in texture-less regions.

\begin{table}[t]
\centering
\scriptsize
\setlength\tabcolsep{2.2pt}
\begin{tabular*}{\columnwidth}{lcC{18pt}C{18pt}C{18pt}C{18pt}C{18pt}C{18pt}c@{}}
	\toprule

  \multirow{3}{*}{Model Architecture} & \multirow{3}{*}{AR} & \multicolumn{3}{c}{Sintel Clean} & \multicolumn{3}{c}{Sintel Final} &  {\multirow{3}{*}{\# Param.}}\\
  \cmidrule(lr){3-5} \cmidrule(lr){6-8}
   & & {ALL} & {NOC} & {OCC} & {ALL} & {NOC} & {OCC} & \\
	\midrule
  {PWC-Net~\cite{sun2018pwc}}                  &   	 & 2.48	& 1.19 & 21.71  & 3.47  & 1.98 & 25.19 &  8.75 M                         \\
  {PWC-Net-small~\cite{sun2018pwc}}            &      & 2.76 & 1.28 & 23.92  & 3.62  & 2.16 & 28.15 & 	4.05 M                     \\ \cmidrule(l{2pt}r{-0pt}){1-9}
  \multirow{2}{*}{{+ Reduce Dense}}   &		   & 2.53	& 1.23 & 21.36  & 3.47  & 2.03 & 24.20 & {\multirow{2}{*}{5.32 M}} \\
                                      &\cmark& 2.04	& 0.90 & 18.47  & 2.97  & 1.72 & 21.05 &                             \\ \cmidrule(l{2pt}r{-0pt}){1-9}
  \multirow{2}{*}{{+ Share Decoder}}  &		   & 2.30 & 1.08 & 20.00  & 3.19  & 1.84 & 22.77 & {\multirow{2}{*}{2.24 M}} \\
                                      &\cmark& 1.95	& 0.85 & 17.85  & 2.86  & 1.66 & 20.25 &                             \\ \cmidrule(l{2pt}r{-0pt}){1-9}
  \multirow{2}{*}{{+ Multipe Frames}} &		   & 2.24 & 1.04 & 19.60  & 3.18  & 1.86 & 22.36 & {\multirow{2}{*}{2.37 M}} \\
                                      &\cmark& 1.89 & 0.86 & 16.79  & 2.85  & 1.66 & 20.02 &\\ \bottomrule
\end{tabular*}
\caption{\textbf{Ablation study of our learning framework with multiple model architectures.} AEPE in specific regions of the scene and the number of CNN parameters are reported.
\textbf{AR}: Training with augmentation as a regularization framework. }
\label{tab:ablation_study_1}
\vspace{-1em}
\end{table}

\subsection{Ablation Study}
To further analyze the capability of each component, we conduct four groups of ablation studies. We randomly re-split the Sintel training set into a new training set and a validation set by the scene.
We evaluate AEPE in different regions over all pixels (ALL), non-occluded pixels (NOC), occluded pixels (OCC), and according to speed (s0-10, s10-40 and s40+ are pixels that move less than 10 pixels, between 10 and 40, and more than 40, respectively)

\begin{table}[t]
\centering
\scriptsize
\setlength\tabcolsep{4.4pt}
\begin{tabular*}{\columnwidth}{@{}llllC{20pt}C{20pt}C{20pt}C{20pt}C{20pt}C{20pt}@{}}
	\toprule
  {} & ST & AT & OT & {ALL} & {NOC} & {OCC} & {s0-10} & {s10-40} & {s40+} \\
	\midrule
	\multirow{6}{*}{{\rotatebox[origin=c]{90}{Sintel Clean}}}
  &	     &    	&   	 & 2.53	& 1.23			& 21.36			& 0.61  & 2.74			& 24.30	\\
  &	     &   	  &\cmark& 2.39	& 1.20			& 19.61			& 0.61  & 2.56			& 23.14 	\\
  &	     &\cmark&		   & 2.40	& 1.13			& 20.67			& 0.61  & 2.81			& 21.95 	\\
  &\cmark&  	  &	     & 2.14	& 1.00			& 18.63			& 0.62  & 2.83			& 17.56 	\\
  &\cmark&\cmark&      & 2.09	& 0.95			& 18.90 		& \bfseries 0.59  & 2.65			& 18.03 	\\
  &\cmark&\cmark&\cmark& \bfseries 2.04	& \bfseries 0.90			& \bfseries  18.47			& 0.61  & \bfseries 2.55			& \bfseries 17.05 	\\ \midrule
	\multirow{6}{*}{{\rotatebox[origin=c]{90}{Sintel Final}}}
  &	     &    	&   	 & 3.47	& 2.03			& 24.20			& 0.82  & 3.77			& 33.48	\\
  &	     &   	  &\cmark& 3.23	& 1.93			& 21.98			& 0.82  & 3.48      & 30.78 	\\
  &	     &\cmark&		   & 3.36	& 1.94			& 23.95			& 0.81  & 3.70      & 32.17	\\
  &\cmark&  	  &	     & 3.04	& 1.78			& 21.25			& 0.78  & 3.55      & 27.80	\\
  &\cmark&\cmark&      & 3.01	& 1.76			& 21.40			& \bfseries 0.75  & 3.48 & 28.48	\\
  &\cmark&\cmark&\cmark& \bfseries 2.97	& \bfseries 1.72 			& \bfseries 21.05			& 0.77  & \bfseries 3.40 & \bfseries 27.25 	\\ \bottomrule
\end{tabular*}
\caption{\textbf{Comparison of combinations of transformations.} AEPE in specific regions are reported. \textbf{ST}: Spatial transformation, \textbf{AT}: Appearance transformation, \textbf{OT}: Occlusion transformation.}
\label{tab:ablation_study_2}
\vspace{-1em}
\end{table}

\vspace{-1em}
\paragraph{Main Ablation.} \cref{tab:ablation_study_1} assesses the overall improvement of our augmentation as a regularization learning framework under multiple model architectures. Our framework consistently improves the accuracy of optical flow over 10\% for all architectures, whether for occluded or non-occluded pixels.

For the consideration of the number of model parameters, we start from the original PWC-Net and a variant named PWC-Net-small without dense connections in the flow decoders~\cite{sun2018pwc}. Although removing dense connections can reduce half parameters, it leads to severe performance degradation. In contrast, our reduced dense variant maintains the performance while reducing 39.2\% parameters. Sharing decoder across feature pyramid yields an improvement on flow with only 25.6\% parameters of the original model. The multi-frame extension reaches the best performance with the minimal extra overhead of parameters.

\vspace{-1em}
\paragraph{Combination of Transformations.} Furthermore, we delve into the type of transformations in our framework.~\cref{tab:ablation_study_2} shows the performance of the model trained with several combinations of the three kinds of transformations. There are some critical observations: \emph{(i)} Each transformation can improve the performance individually. \emph{(ii)} Spatial transformation is the most helpful to all measurements, especially for large displacement estimation. \emph{(iii)} The accuracy in the occluded region can be significantly improved by occlusion transformation or spatial transformation. All these observations are consistent with our assumption that the transformation will introduce new challenging scenes, and our approach can provide reliable supervision.

\begin{table}[t]
\centering
\scriptsize
\setlength\tabcolsep{3.5pt}
\begin{tabular*}{\columnwidth}{@{}lcccccccc@{}}
	\toprule
	\multirow{3}{*}{Method} & \multicolumn{4}{c}{Sintel Clean} & \multicolumn{4}{c}{Sintel Final}\\
	\cmidrule(lr){2-5} \cmidrule(lr){6-9}
	                        & ALL & s0-10 & s10-40 & s40+ & ALL & s0-10 & s10-40 & s40+\\

    \midrule
  Without Aug.       & 2.53 & \bfseries0.61 & 2.74 & 24.30 & 3.47 & 0.82 & 3.77 & 33.48 \\
	Aug. Directly 	   & 2.71 & 0.69 & 3.11 & 27.13 & 3.80 & 0.95 & 4.03 & 35.90\\
	Aug. Distillation  & 2.36 & 0.64 & 2.61 & 19.90 & 3.31 & 0.86 & 3.50 & 30.18 \\
  Ours(aug. as reg.) & \bfseries2.04 & \bfseries0.61 & \bfseries2.55 & \bfseries17.05 &  \bfseries2.97 & \bfseries0.77 & \bfseries3.40 & \bfseries27.25    \\ \bottomrule
\end{tabular*}
\caption{Comparison of our learning framework with direct data augmentation and the data distillation framework used in~\cite{liu2019ddflow, liu2019selflow}.}
\label{tab:usage}
\vspace{-1em}
\end{table}

\begin{table}[t]
\centering
\scriptsize
\setlength\tabcolsep{3.5pt}
\begin{tabular*}{\columnwidth}{@{}lcccccccc@{}}
	\toprule
	\multirow{3}{*}{Method} & \multicolumn{4}{c}{Sintel Clean} & \multicolumn{4}{c}{Sintel Final}\\
	\cmidrule(lr){2-5} \cmidrule(lr){6-9}
	                        & ALL & s0-10 & s10-40 & s40+ & ALL & s0-10 & s10-40 & s40+\\
    \midrule
	Without Aug.       & 2.53 & 0.61 & 2.74 & 24.30 & 3.47 & 0.82 & 3.77 & 33.48 \\
	CPAB~\cite{freifeld2017cpa} + AT   & 2.38 & 0.61 & 2.78 & 21.60 & 3.32 & 0.81 & 3.59 & 31.09\\
	AutoAugment~\cite{cubuk2019autoaugment}& 2.30 & 0.62 & 2.59 & 21.18 & 3.29 & 0.81 & 3.53 & 30.11 \\
  Ours(ST + AT) & \bfseries2.09 & \bfseries0.59 & \bfseries2.65 & \bfseries18.03 &  \bfseries3.01 & \bfseries0.75 & \bfseries3.48 & \bfseries28.48    \\ \bottomrule
\end{tabular*}
\caption{Comparison of different augmentation transformations integrated with our framework. AT: appearance transformation, ST: spatial transformation.}
\label{tab:complicate_aug}
\vspace{-10pt}
\end{table}

\vspace{-1em}
\paragraph{Usage of Augmentation.}
As we mentioned above, almost all of the unsupervised learning approaches avoid using a heavy combination of augmentations. As a reference, we evaluate the same transformations with different usages. ~\cref{tab:usage} reports the results of \emph{(i)} training without heavy augmentation, \emph{(ii)} using transformation as a regular data augmentation and training directly, \emph{(iii)} training with data distillation that similar in~\cite{liu2019ddflow, liu2019selflow}, \emph{(iv)} training with the learning framework we proposed. The results show that directly augmentation makes all metrics worse. Instead of applying transformations directly, distillation alleviates the problem of unreliable supervision. However, the frozen teacher model is still a bottleneck for the student model. Also, the tedious multi-stage training process of knowledge distillation is undesired. Our framework avoids the unreliable photometric loss for the transformed samples. It achieves the best results with a single-stage optimization.

\vspace{-1em}

\paragraph{Integrate Complicated Augmentation.}
By implementing the corresponding transformation of optical flow and occlusion map, our framework can be integrated with almost all types of augmentation. We assess a complicated spatial transformation called CPAB~\cite{freifeld2017cpa} and a recent work in AutoML on searching for the best augmentation policy called AutoAugment~\cite{cubuk2019autoaugment}. Note that random zoom in is applied first to avoid invalid coordinate values of transformations. \cref{tab:complicate_aug} shows that both strategies integrated with our framework can improve accuracy. Note that AutoAugment is too time consuming for our task, therefore we adopt the final policy searched from ImageNet~\cite{imagenet_cvpr09} classification task.
It is promising that our framework with AutoAugment will be further improved with policy fine-tuning.

\begin{table}[t]
\centering
\scriptsize
\setlength\tabcolsep{4.3pt}
\begin{tabular*}{\columnwidth}{@{}lC{1pt}cC{1pt}ccccc@{}}
	\toprule
	\multirow{2}{*}{Method} && Training & & Chairs & Sintel & Sintel & KIITI & KITTI \\
  && Set & & Full & Clean & Final & 2012 & 2015 \\

    \midrule
	\multirow{1}{*}{PWC-Net~\cite{sun2018pwc}}
	&& Sintel && 3.69 & \bfseries(1.86) & \bfseries(2.31) & 3.68  & 10.52\\ \midrule
	\multirow{2}{*}{Ours(ARFlow)}
  && Sintel && \bfseries3.50 & (2.79) & (3.73) & 3.06 & 9.04 \\
	&& CityScapes && 5.10 & 5.22 & 6.01 & \bfseries2.11  & \bfseries5.33 \\  \bottomrule
\end{tabular*}
\caption{Generalization performance of cross datasets evaluation. The numbers indicate AEPE on each dataset. For KITTI and Sintel, the results are evaluated on the training set. () indicates the results of a dataset that the method has been trained on.}
\label{tab:cross}
\vspace{-10pt}
\end{table}

\vspace{-.1em}
\subsection{Cross Dataset Generalization}
Although deep optical flow methods have been far ahead of the most popular classical variational method TV-L1~\cite{wedel2008tvl1} on optical flow benchmarks, the latter has not gone away. One possible reason is that supervised learning methods are prone to overfitting, which results in poor generalization when transferring to high-level video tasks.

Hence, we report the cross dataset accuracy in~\cref{tab:cross}, in which our unsupervised method is compared with a fully supervised method PWC-Net~\cite{sun2018pwc}. The supervised PWC-Net consistently outperforms for the dataset that the model is trained on, while our unsupervised method works much better when transferring to other datasets. In addition, we train a model on an urban street dataset named CityScapes~\cite{Cordts2016Cityscapes}, in which 50,625 image pairs are used for training without the ground truth. This model performs best on the KITTI 2012 and KITTI 2015 than any other model trained on the synthetic dataset. Our method makes it possible to fit the domain of high-level video tasks by training a model on the unlabeled videos from that domain.

Remarkably, despite the lack of cross dataset results from other unsupervised methods, the accuracy of our model trained on CityScapes is even better than most of the previous works trained on KITTI (\cf~\cref{tab:flow_KITTI}), which shows the superiority of our method. The results demonstrate a significant improvement of our method for unsupervised optical flow task with an excellent generalization.

\vspace{-.3em}
\section{Conclusion}
We proposed a novel framework that learns optical flow from unlabeled image sequences with the self-supervision from augmentations. To avoid the objective of view synthesis being unreliable on transformed data, we twist the basic learning framework by adding another forward pass for transformed images, where the supervision is from the transformed prediction of original images. Besides, a lightweight network and its multi-frame extension were presented. Extensive experiments have shown that our methods significantly improve accuracy, with high compatibility and generalization ability. We believe that our learning framework can be further combined with other geometrical constraints or transferred to other visual geometry tasks, such as depth or scene flow estimation.

\myparagraph{Acknowledgment} We thank anonymous reviewers for their constructive comments, and LL would like to thank Pengpeng Liu for helpful suggestions. This work is partially supported by the National Natural Science Foundation of China (NSFC) under Grant No. 61836015 and Key R\&D Program Project of Zhejiang Province (2019C01004).

{\small
\bibliographystyle{ieee_fullname}
\bibliography{ref}
}

\end{document}